\documentclass[sigconf]{acmart}
\AtBeginDocument{%
  }
\usepackage{caption}
\usepackage{tabularx}
\usepackage{subcaption}
\setcopyright{acmlicensed}
\copyrightyear{2026}
\acmYear{2026}
\acmDOI{}
\acmConference[D-TUR '26]{HRI '26 Designing Transparent and Understandable Robots}{March 16,
  2026}{Edinburgh, UK}
\acmISBN{}

\begin{document}

\title{Dance2Hesitate: A Multi-Modal Dataset of Dancer-Taught Hesitancy for Understandable Robot Motion}

\author{Srikrishna Bangalore Raghu}
\affiliation{%
 \institution{University of Colorado Boulder}
 \city{Boulder}
 \state{Colorado}
 \country{United States}
}\email{srikrishna.bangaloreraghu@colorado.edu}

\author{Anna Soukhovei}
\affiliation{%
 \institution{University of Colorado Boulder}
 \city{Boulder}
 \state{Colorado}
 \country{United States}
}\email{anna.soukhovei@colorado.edu}

\author{Divya Sai Sindhuja Vankineni}
\affiliation{%
 \institution{University of Colorado Boulder}
 \city{Boulder}
 \state{Colorado}
 \country{United States}
}\email{divya.vankineni@colorado.edu}

\author{Alexandra Bacula}
\affiliation{%
 \institution{Pacific Lutheran University }
 \city{Tacoma}
 \state{Washington}
 \country{United States}
}\email{alexandra.bacula@plu.edu}

\author{Alessandro Roncone}
\affiliation{%
 \institution{University of Colorado Boulder}
 \city{Boulder}
 \state{Colorado}
 \country{United States}
}\email{alessandro.roncone@colorado.edu}



    
\begin{abstract}
In human-robot collaboration, a robot's expression of \textsl{hesitancy} is a critical factor that shapes human coordination strategies, attention allocation, and safety-related judgments. However, designing \textsl{hesitant} robot motion that generalizes is challenging because the observer's inference is highly dependent on embodiment and context. To address these challenges, we introduce and open-source a multi-modal, dancer-generated dataset of \textsl{hesitant} motion where we focus on specific context-embodiment pairs (i.e., manipulator/ human upper-limb approaching a Jenga Tower, and anthropomorphic whole body motion in free space). The dataset includes (i) kinesthetic teaching demonstrations on a Franka Emika Panda reaching from a fixed start configuration to a fixed target (a Jenga tower) with three graded hesitancy levels (slight, significant, extreme) and (ii) synchronized RGB-D motion capture of dancers performing the same reaching behavior using their upper limb across three hesitancy levels, plus full human body sequences for extreme hesitancy. We further provide documentation to enable reproducible benchmarking across robot and human modalities. Across all dancers, we obtained 70 unique whole-
body trajectories, 84 upper limb trajectories spanning over the three
hesitancy levels, and 66 kinesthetic teaching trajectories spanning
over the three hesitancy levels. The dataset can be accessed here: https://brsrikrishna.github.io/Dance2Hesitate/. 
\end{abstract}

\ccsdesc[500]{Human-centered computing~Empirical studies in HCI}
\ccsdesc[500]{Human-centered computing~Empirical studies in interaction design}
\keywords{Human-Robot Interaction, Expressive Motion, Hesitancy }


\renewcommand{\shortauthors}{Bangalore Raghu et al.}
\maketitle

\begin{figure}[!t]
    \centering
  \includegraphics[width=\linewidth]{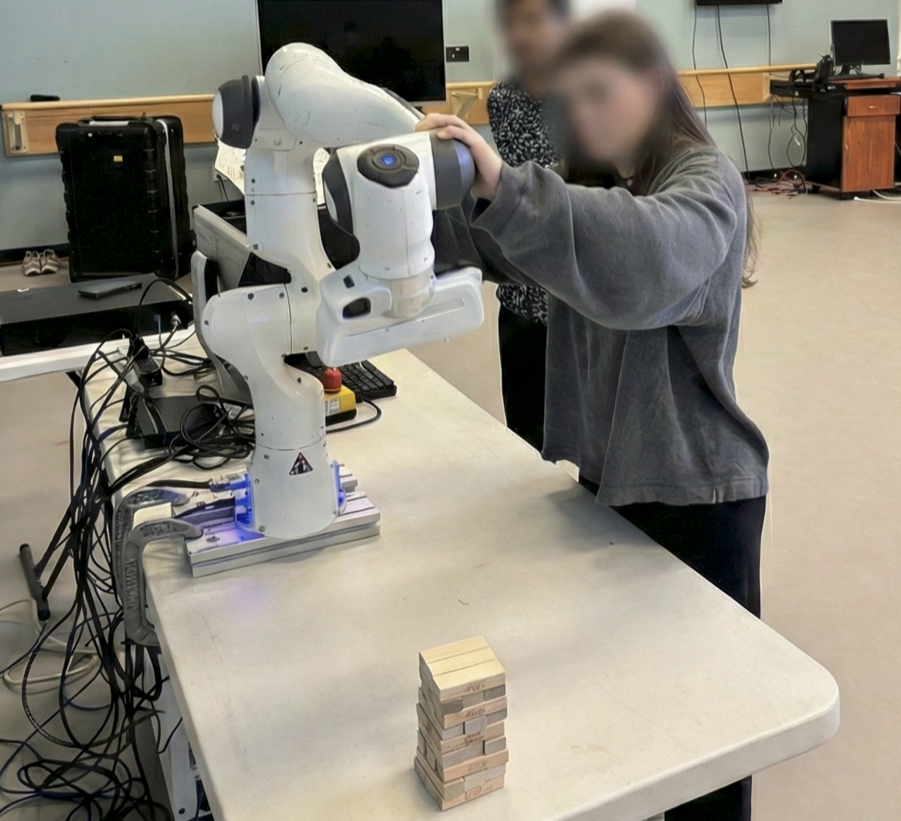}
  \caption{A dancer performing kinesthetic teaching on the Franka Emika Panda manipulator to generate a \textsl{hesitant} trajectory towards the Jenga tower}.
  \label{fig:kinesthetic}
\end{figure}
\section{Introduction}

As robots become more widespread in our everyday lives, ensuring successful collaboration with humans has become exceedingly important \cite{semeraro2023human}. Coherent human--robot collaboration requires the robot to transparently communicate its intention and sentiment, while the human attempts to form an accurate mental model of the robot (i.e., an understanding of its capabilities, intentions, and limitations) \cite{tabrez2020mentalmodels}. While speech is a dominant channel for conveying information, humans regularly complement verbal communication with nonverbal cues \cite{knapp2021nonverbalcomm}. Employing robots that exhibit similar nonverbal signals can enhance coordination, cohesion, and shared understanding within human--robot teams \cite{fong2003socialrobots,admoni2016robot_nonverbal_behavior, breazeal2005nonverbal_teamwork}. Nonverbal communication is inherently multimodal, encompassing gaze, haptics, facial expressions, and other bodily cues \cite{knapp2021nonverbalcomm}. In robotics, a widely applicable modality is motion, where spatiotemporal variations in the robot joints can serve as readable signals of its internal state \cite{hoffmanju2014movement,dragan2013legibility}.\\
Expressive robot motion has been studied extensively in the context of affective expression, where robots convey emotions such as happiness, sadness, or fear through stylistic variations in movement \cite{knight2014laban,suguitan2020moveae}. In this work, we focus on \textsl{hesitancy}, which falls under the umbrella of \textsl{functional expressivity} (i.e., the robot’s ability to communicate its functional capabilities and limitations) \cite{raghu2025employing}. In pursuit of exploring this direction of research, we frame ``uncertainty'' as the functional capability of the robot to express the unreliability and riskiness of successfully completing its task, and \textsl{hesitancy} as the functional expression, i.e., what is used to express uncertainty. \textsl{Hesitancy} is particularly consequential in human--robot collaboration since it directly shapes how humans coordinate with the robot, readjust their own attention, and make safety-critical judgments \cite{dragan2015effects,hough2017uncertainty}. When a robot is uncertain about its success, conveying it through motion can help humans calibrate trust, anticipate potential failure, and intervene appropriately, thereby reducing unsafe surprises and improving teamwork \cite{wang2016trust,moon2013ahp}.\\
However, achieving this is extremely challenging due to two major reasons. Due to regular interactions between humans in everyday life, we tend to relate to a humanoid robot’s form factor and easily interpret its nonverbal cues \cite{fink2012anthropomorphism}, which does not translate to most robots with non-anthropomorphic morphologies (e.g., manipulators, mobile bases, or UAVs). As a result, the same underlying internal state can produce motion cues that are ambiguous or even misinterpreted when expressed through different embodiments, since each platform offers different degrees of freedom and kinematic constraints \cite{hoffmanju2014movement,fong2003socialrobots}. Second, expressive motion is inherently context-specific: how a behavior such as hesitancy is perceived depends strongly on the task, environment, and interaction dynamics \cite{moon2013ahp,hough2017uncertainty}. For example, a pause or a slow approach may signal caution, uncertainty, or deference in one setting, but may be interpreted as malfunction, inefficiency, or obstruction in another \cite{moon2013ahp,hoffmanju2014movement}. \\Together, these embodiment and context dependencies make it difficult to design expressive behaviors that generalize beyond a single robot or scenario, motivating controlled experimental tasks and multi-modal datasets that isolate expressive intent while holding the functional goal constant. In this work, we take a deliberate first step toward addressing this challenge by studying hesitancy in two carefully chosen context-embodiment pairs that emphasize both task-grounded and embodiment-grounded expression: (i) a fixed-goal manipulation setting in which a Franka Emika Panda approaches a Jenga tower from a canonical start pose, and (ii) anthropomorphic whole-body motion in free space that captures broader posture and weight-shift cues. We position this release as a controlled first step that establishes reproducible benchmarks and a cross-modal structure that future work can extend to additional robot platforms and scenarios. To achieve this, we recruited dancers to ensure reliable, repeatable, and intentionally modulated demonstrations of graded hesitancy under a fixed task context. Dancers’ movement expertise helps reduce unwanted variability and enables clearer separation between hesitancy levels, which is valuable for establishing a benchmark and extracting kinematic signatures of hesitancy, resulting in the following contributions:
\begin{enumerate}
    \item \textbf{A multi-modal, dancer-generated dataset of hesitant motion} that captures functional expressivity across three modalities of data, consisting of kinesthetic teaching demonstrations on a Franka Emika Panda reaching from a fixed start configuration to a fixed target (a Jenga tower) with three graded hesitancy levels (slight, significant, extreme), alongside RGB-D motion capture of dancers performing the same reaching behavior with upper-limb keypoints at all three levels and full human-body sequences for extreme hesitancy.
    \item \textbf{A unified data release with standardized formats and tooling}, including MP4 videos, ROSBags, CSV/NPZ robot logs, synchronized RGB videos, and 2D/3D keypoints with confidence scores, enabling reproducible benchmarking across robot and human modalities.
\end{enumerate}

\section{Related Work}
\label{sec:related_work}

\newcommand{\img}[1]{\includegraphics[width=0.18\textwidth]{#1}}

\begin{figure*}[t]
\centering
\setlength{\tabcolsep}{2pt}
\renewcommand{\arraystretch}{0}

\begin{tabular}{@{}c c c c c c@{}}
& \scriptsize (1) & \scriptsize (2) & \scriptsize (3) & \scriptsize (4) & \scriptsize (5) \\[2pt]

\rotatebox{90}{\scriptsize Upper-limb (front)} &
\img{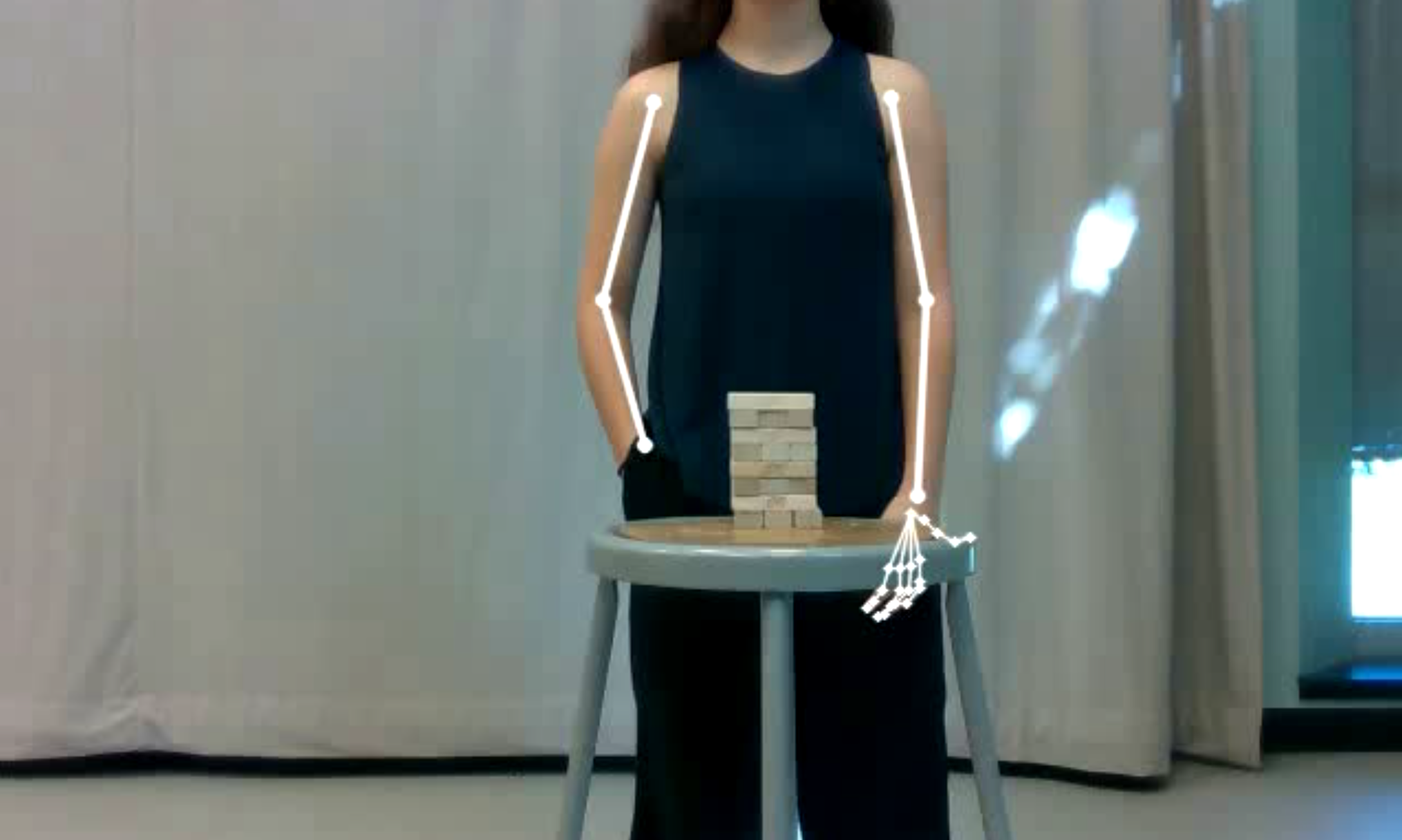} &
\img{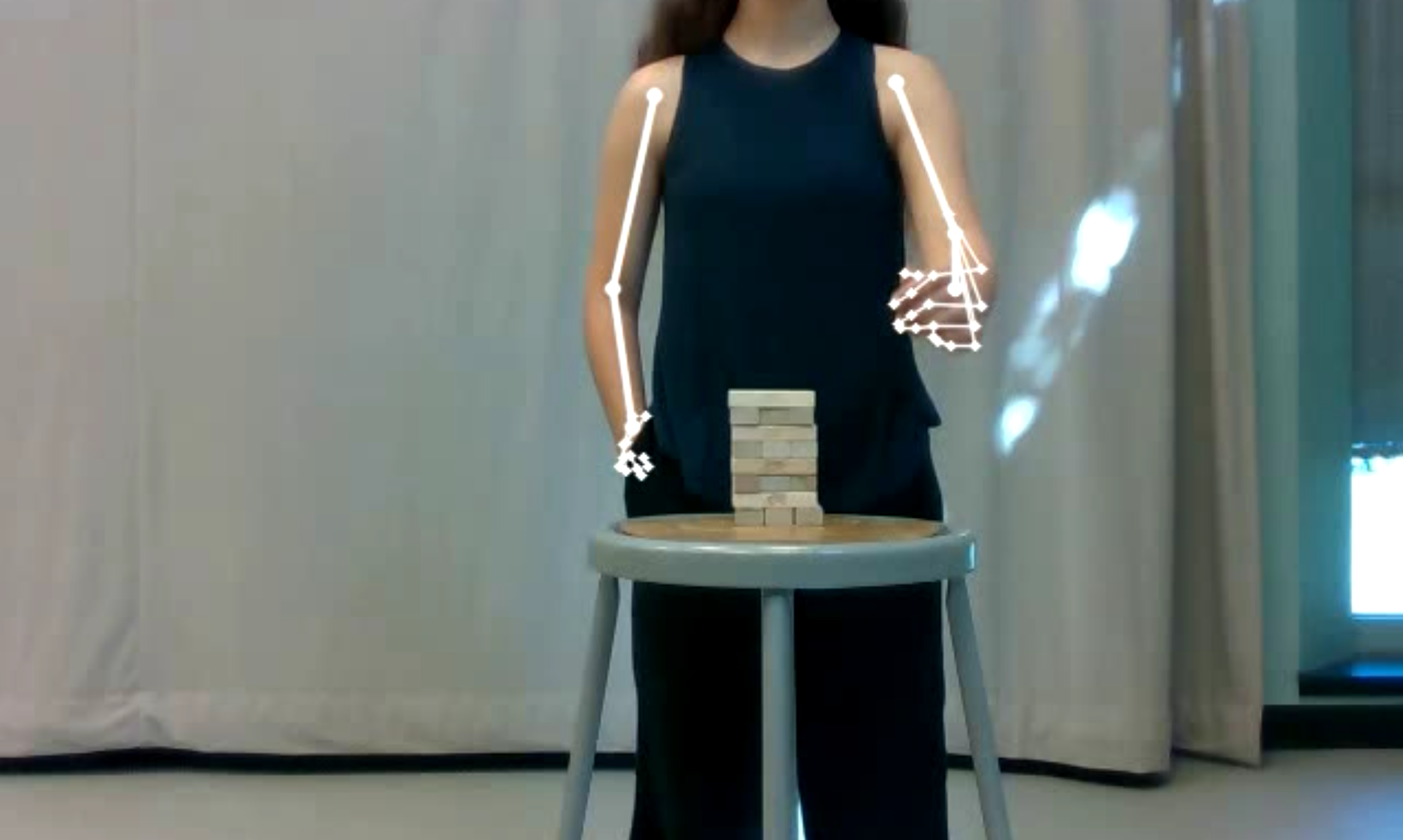} &
\img{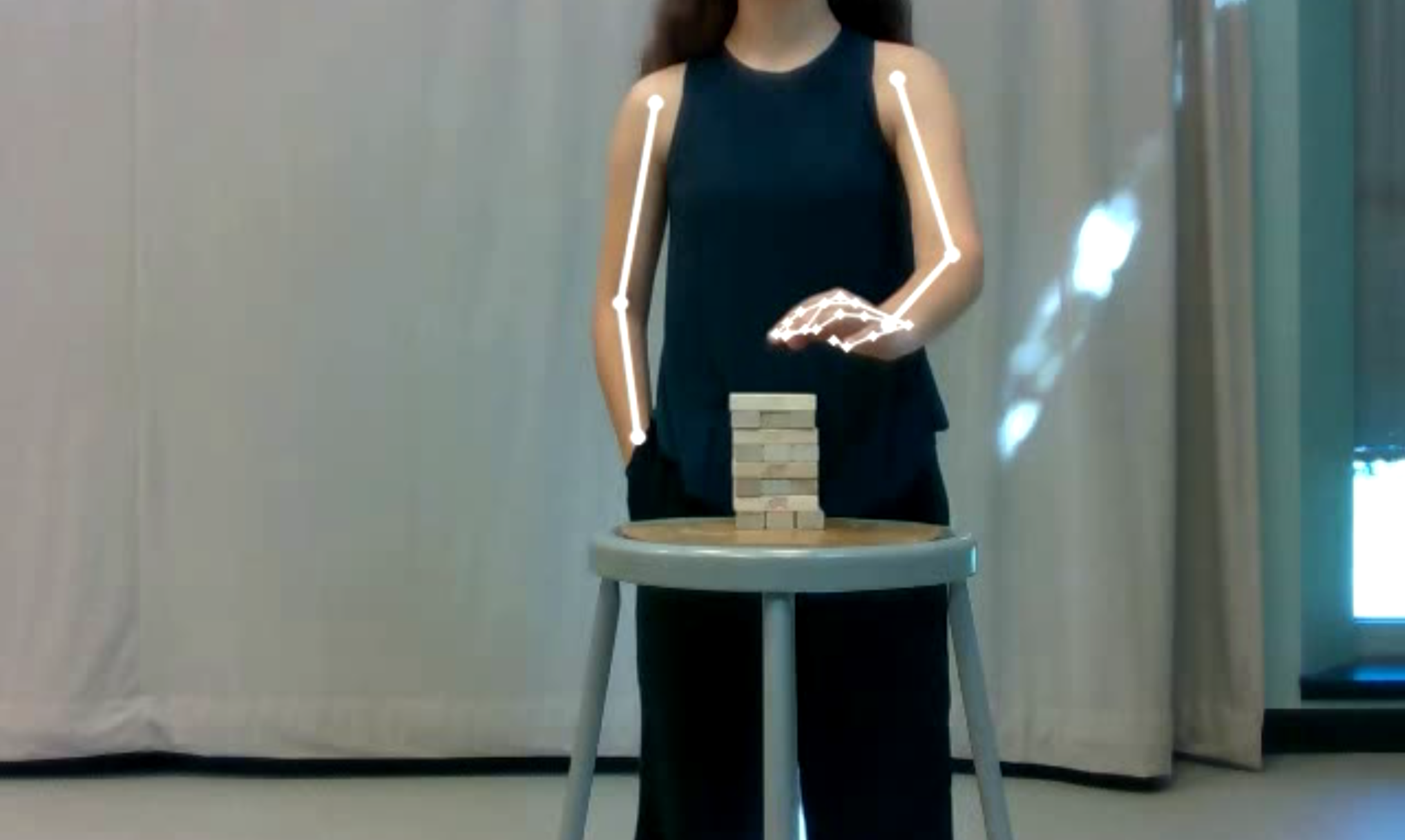} &
\img{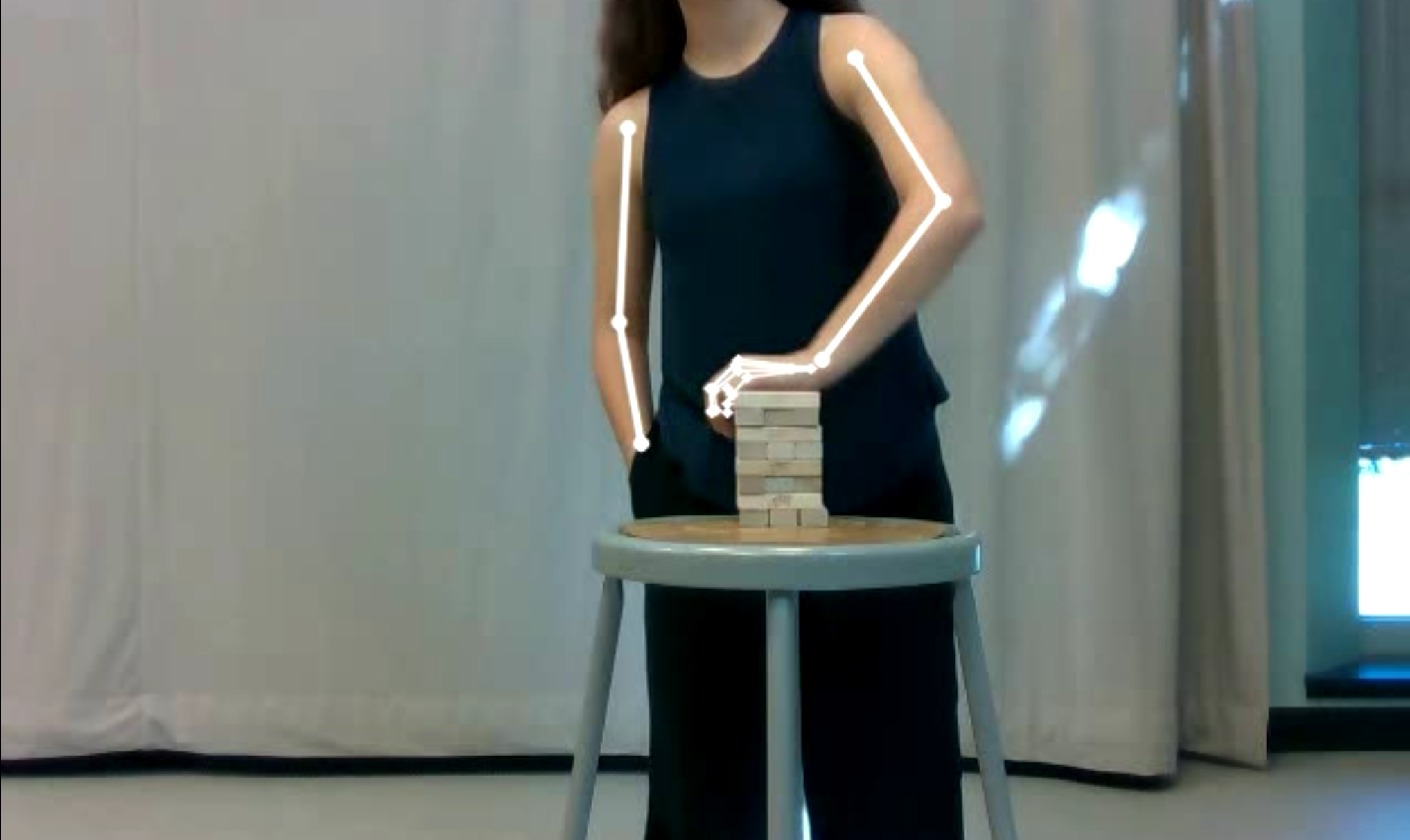} &
\img{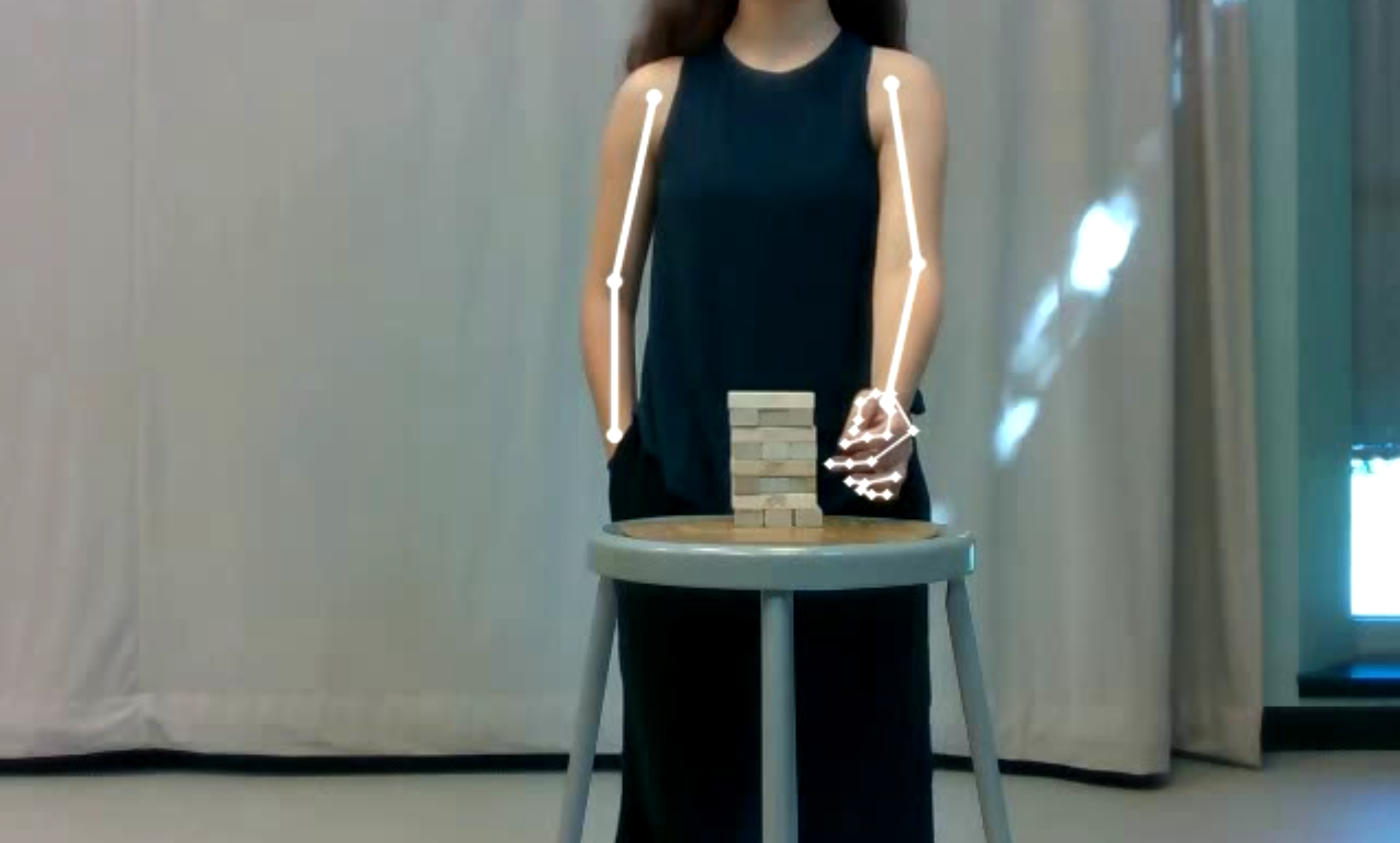} \\[2pt]

\rotatebox{90}{\scriptsize Upper-limb (side)} &
\img{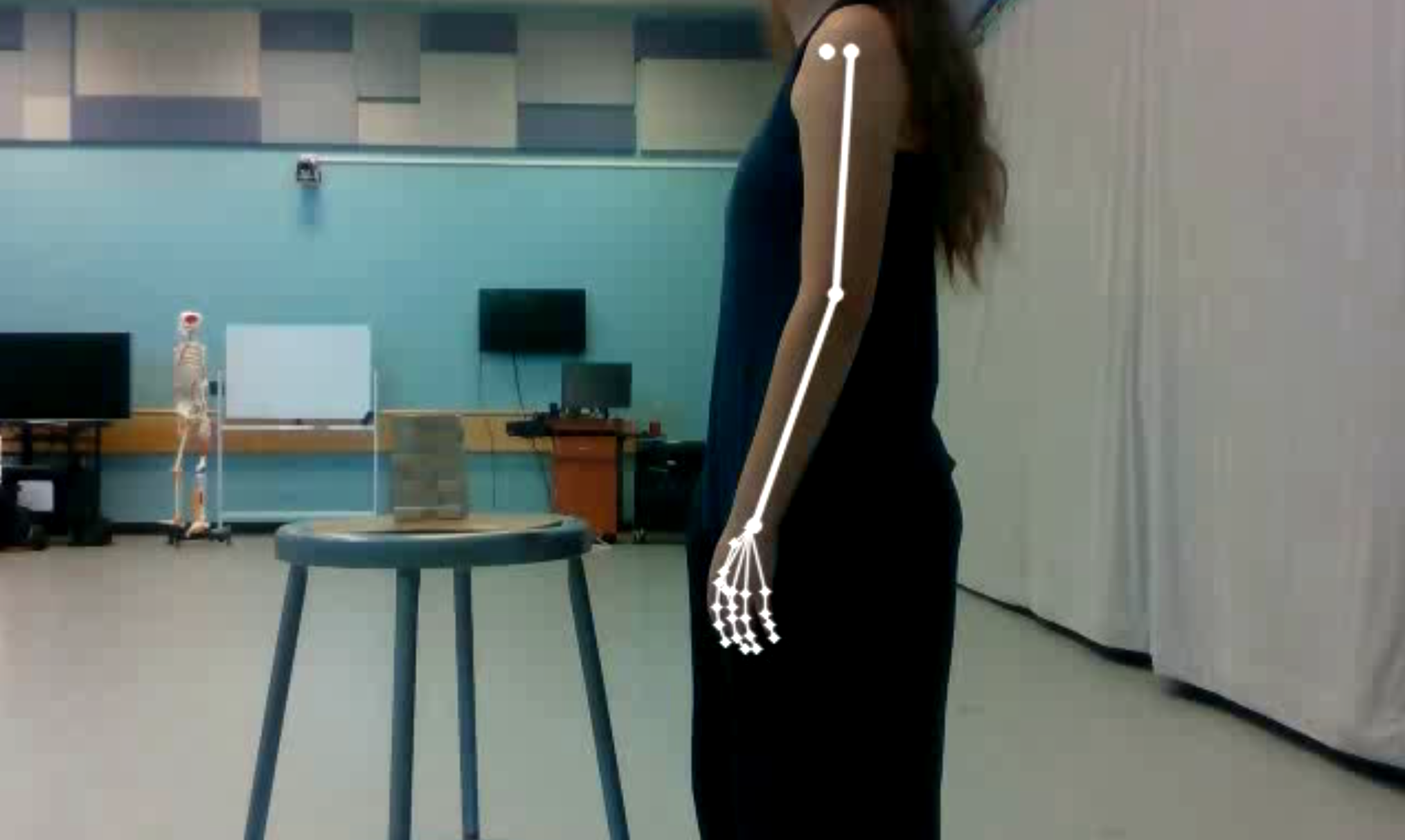} &
\img{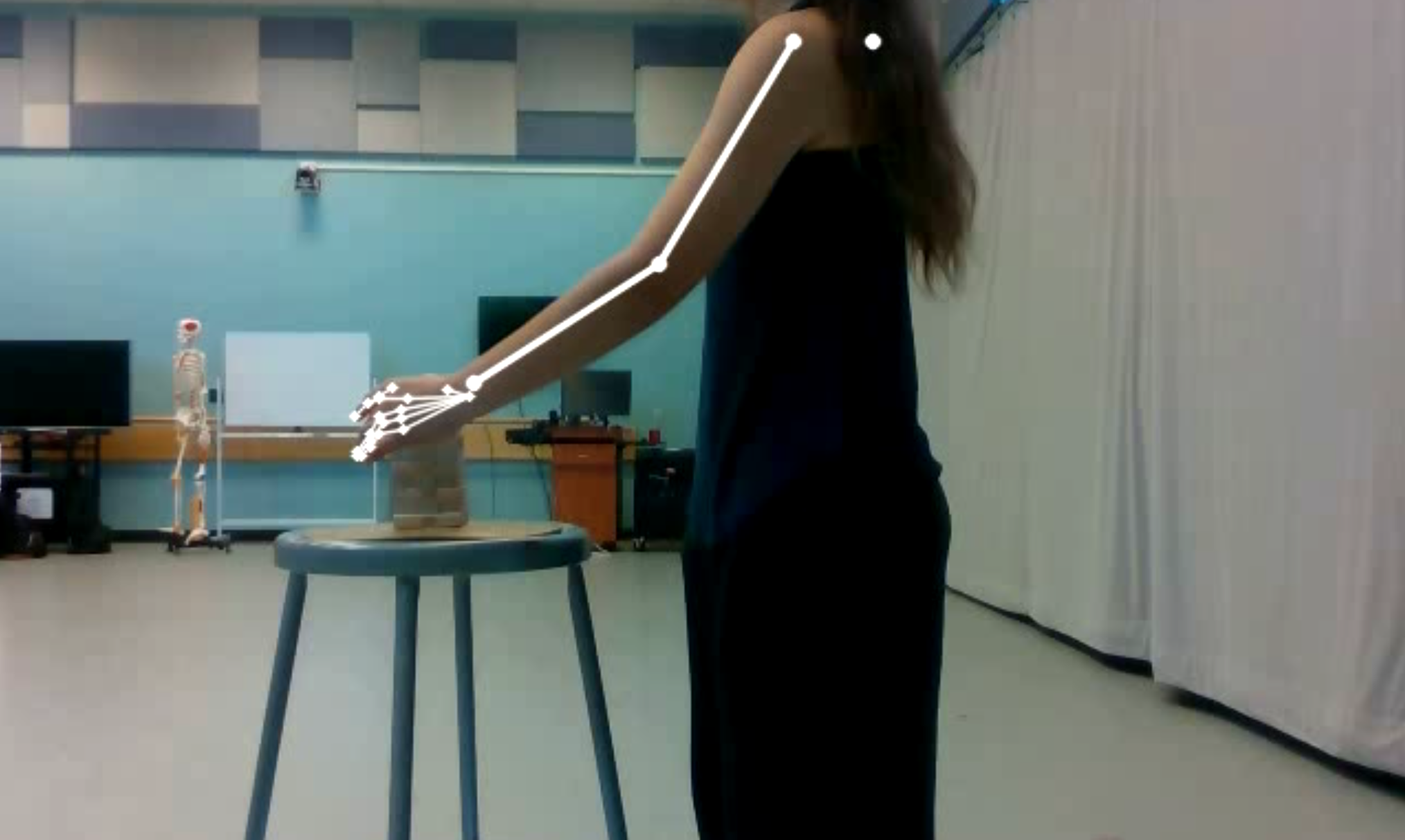} &
\img{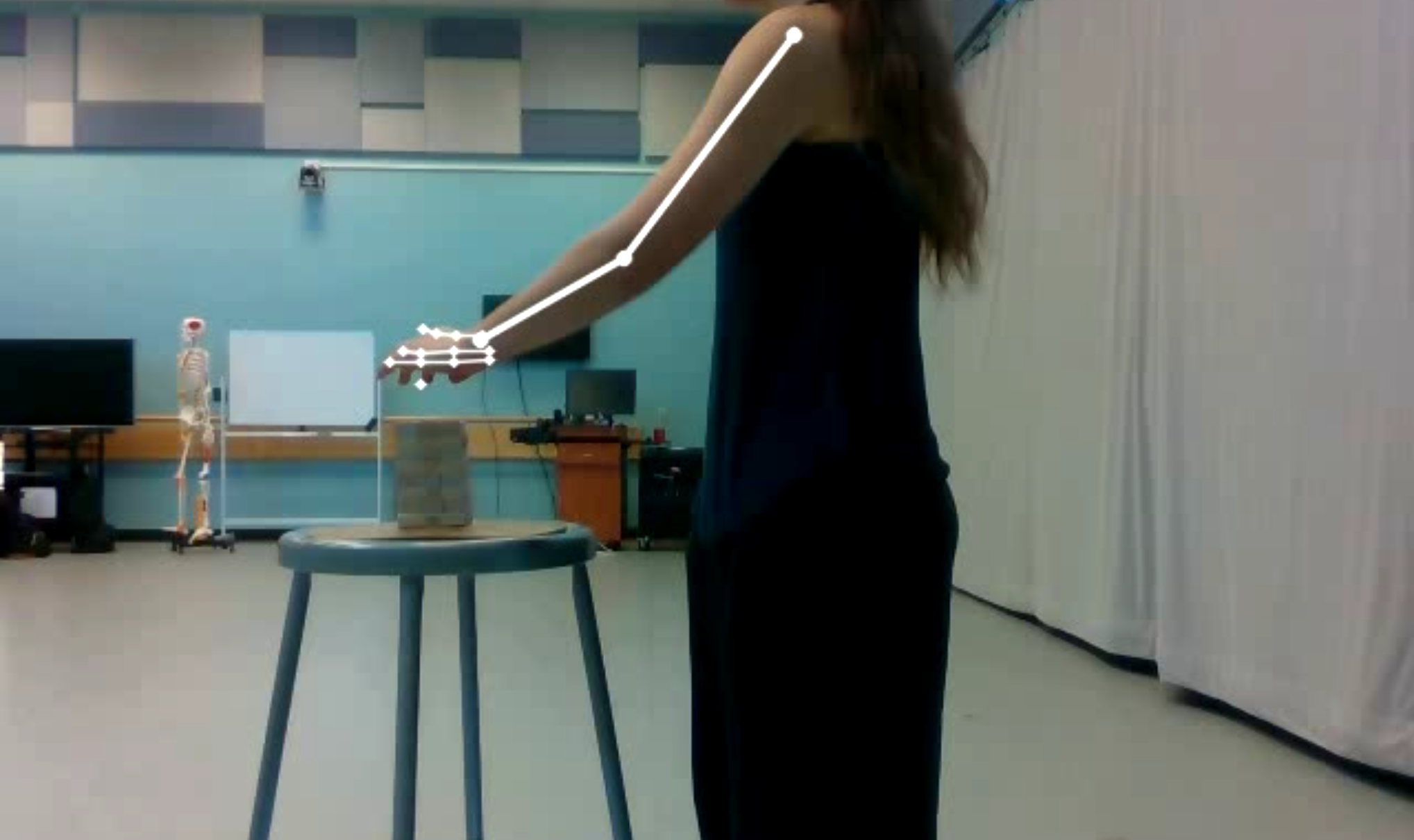} &
\img{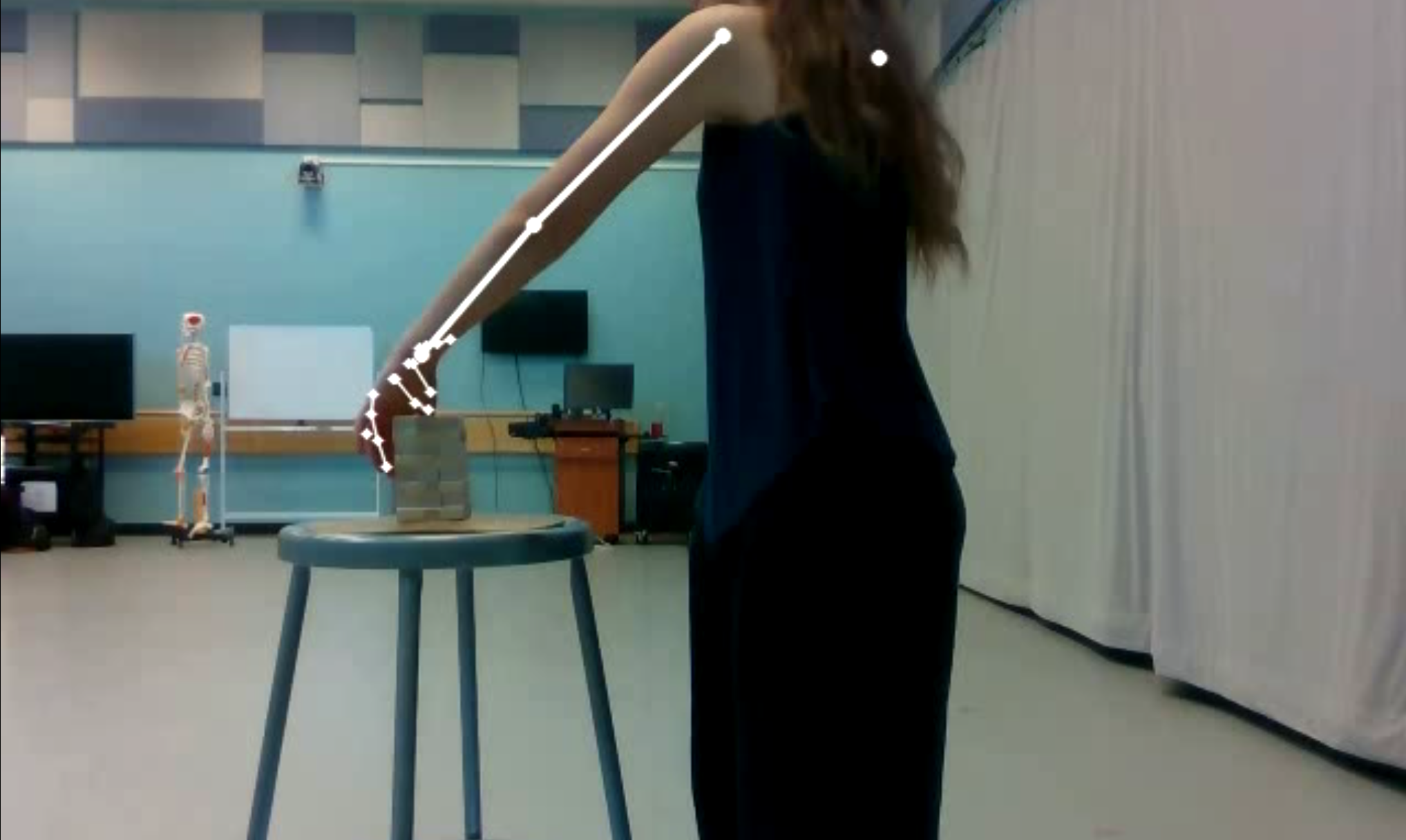} &
\img{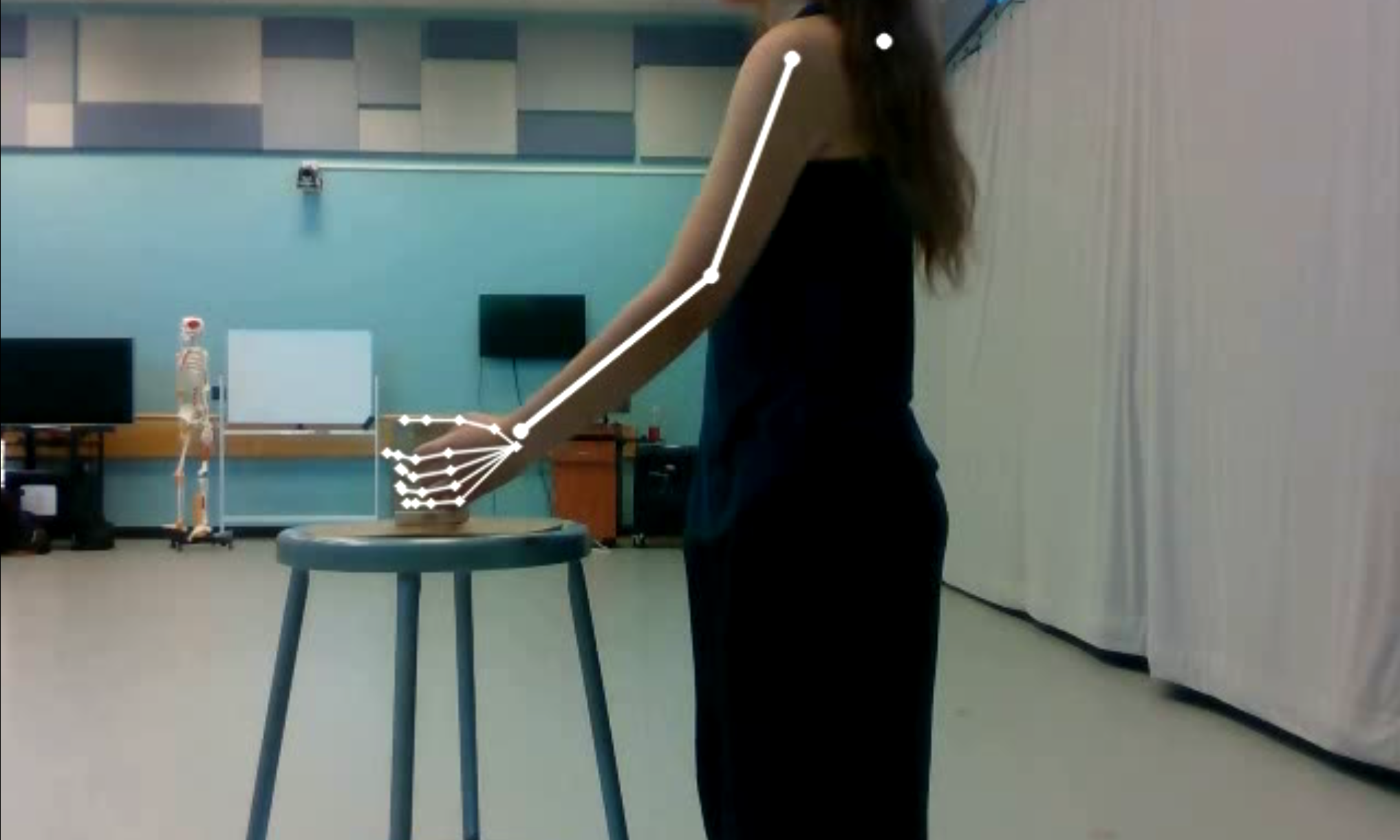} \\[2pt]

\rotatebox{90}{\scriptsize Whole-body (front)} &
\img{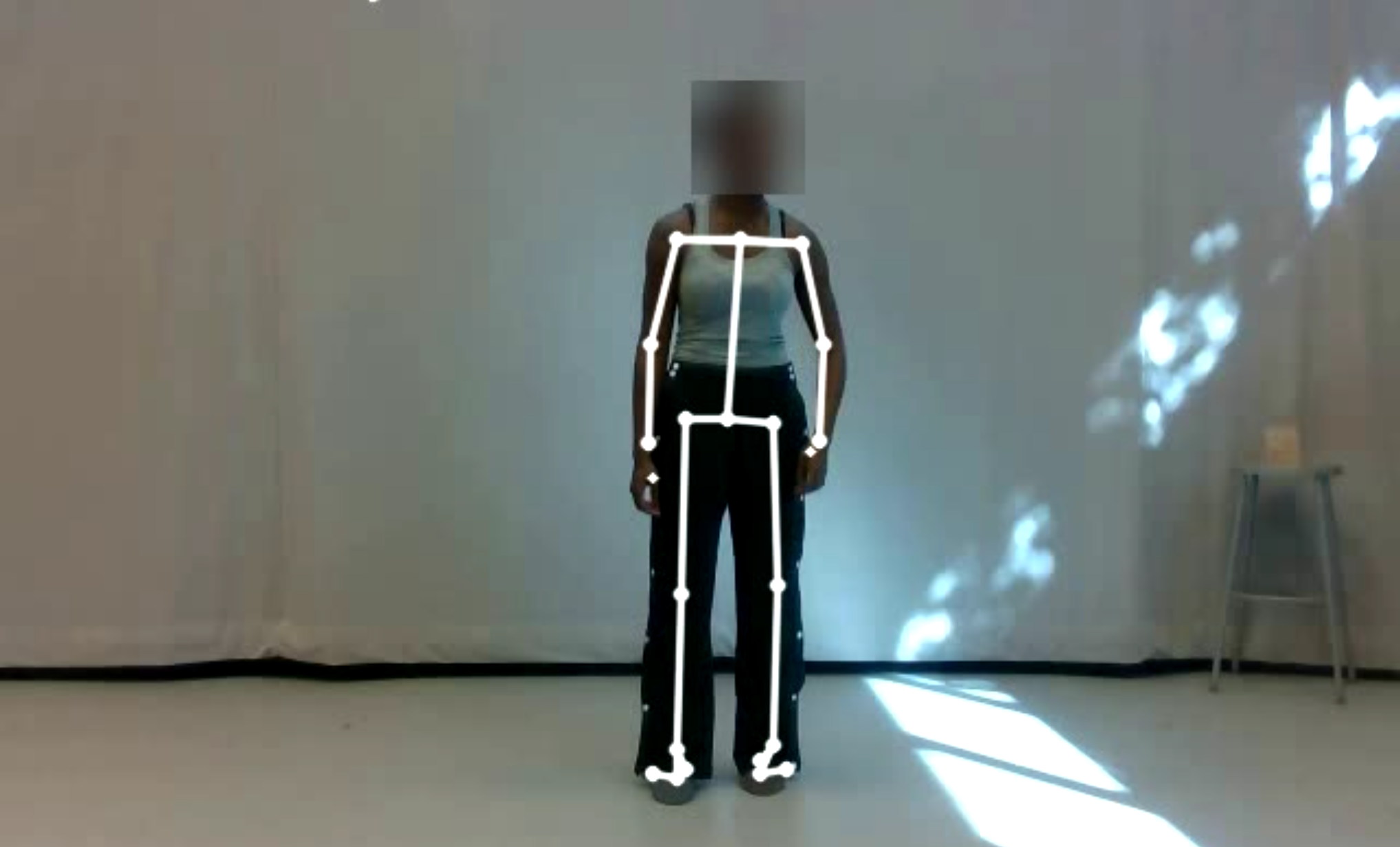} &
\img{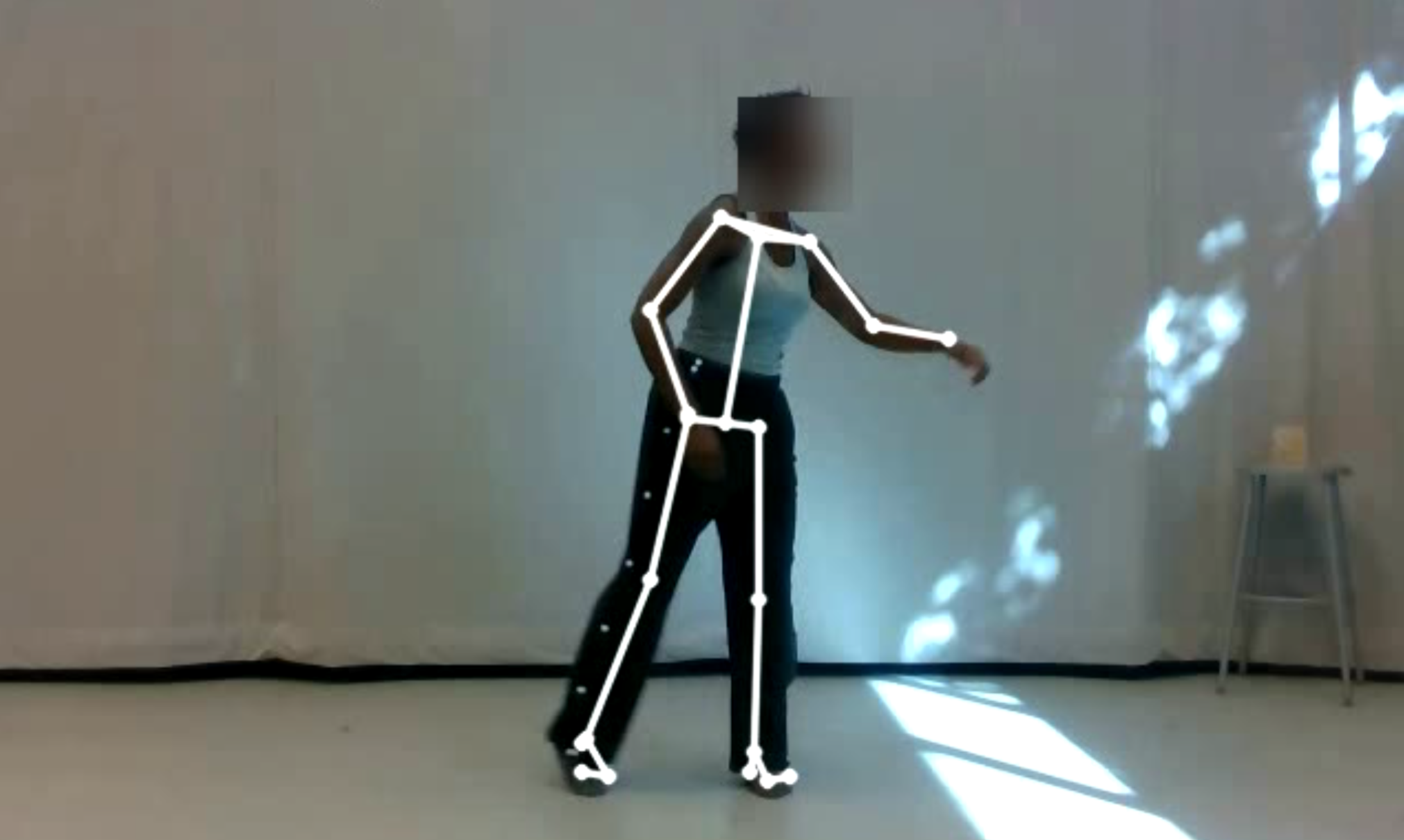} &
\img{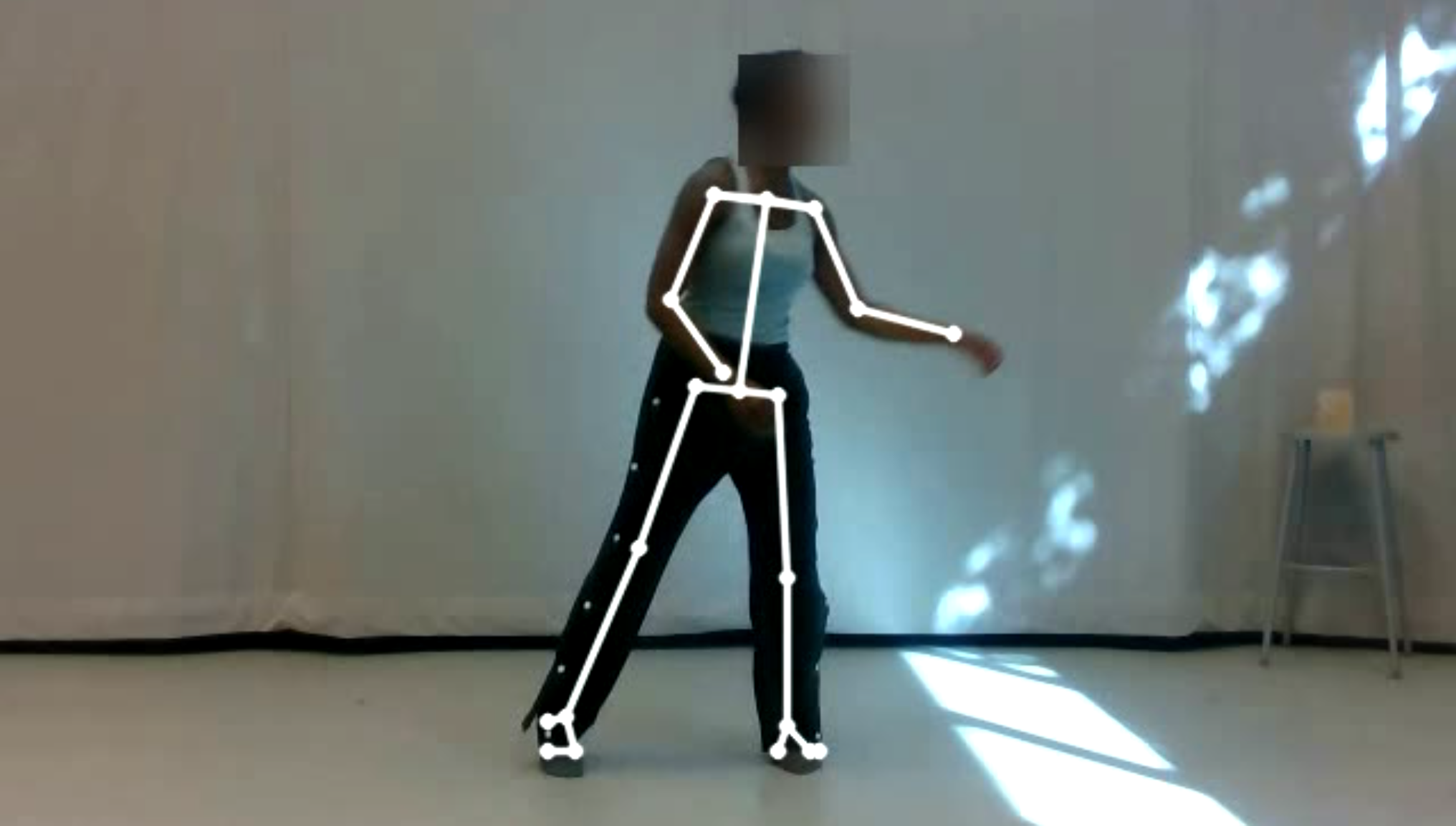} &
\img{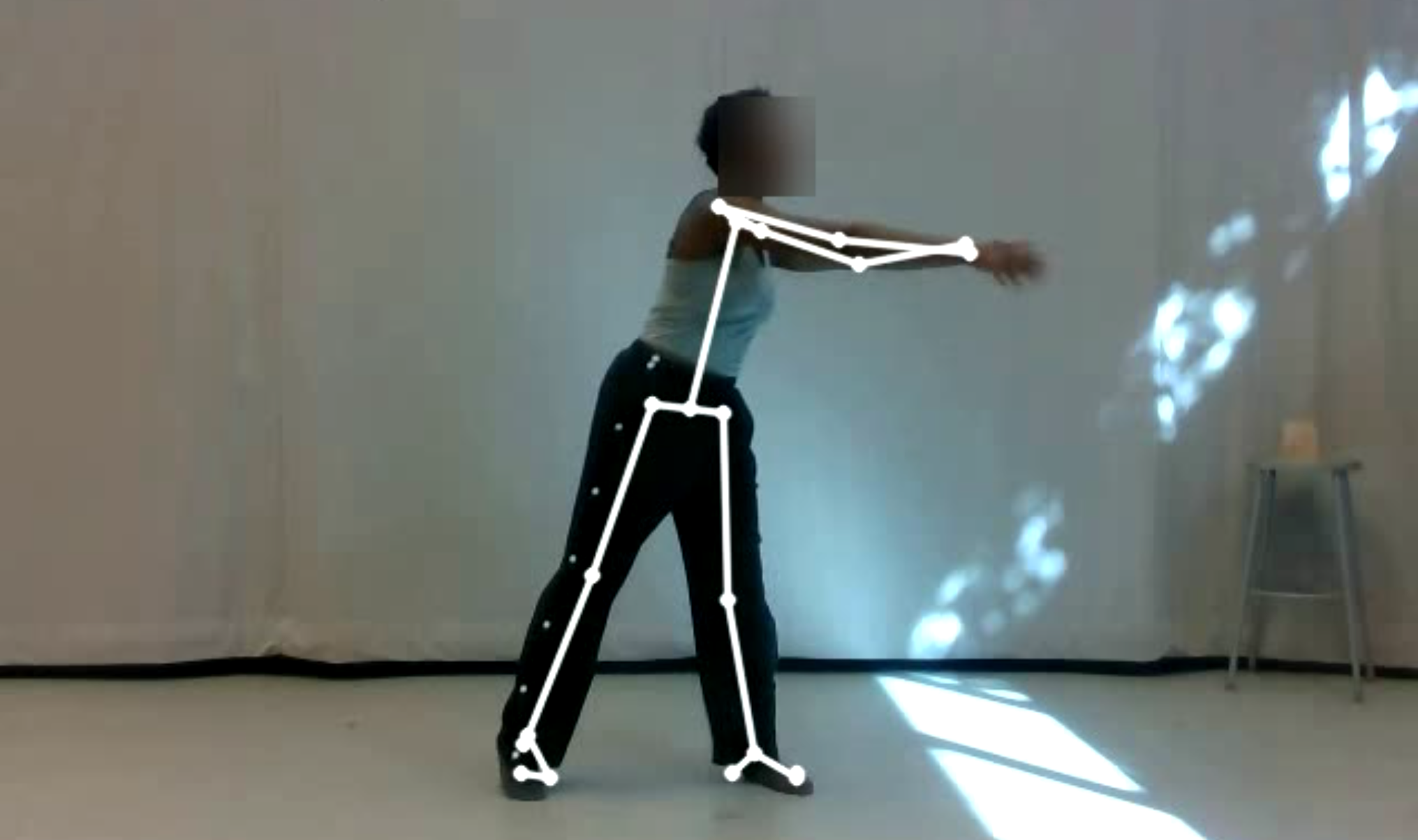} &
\img{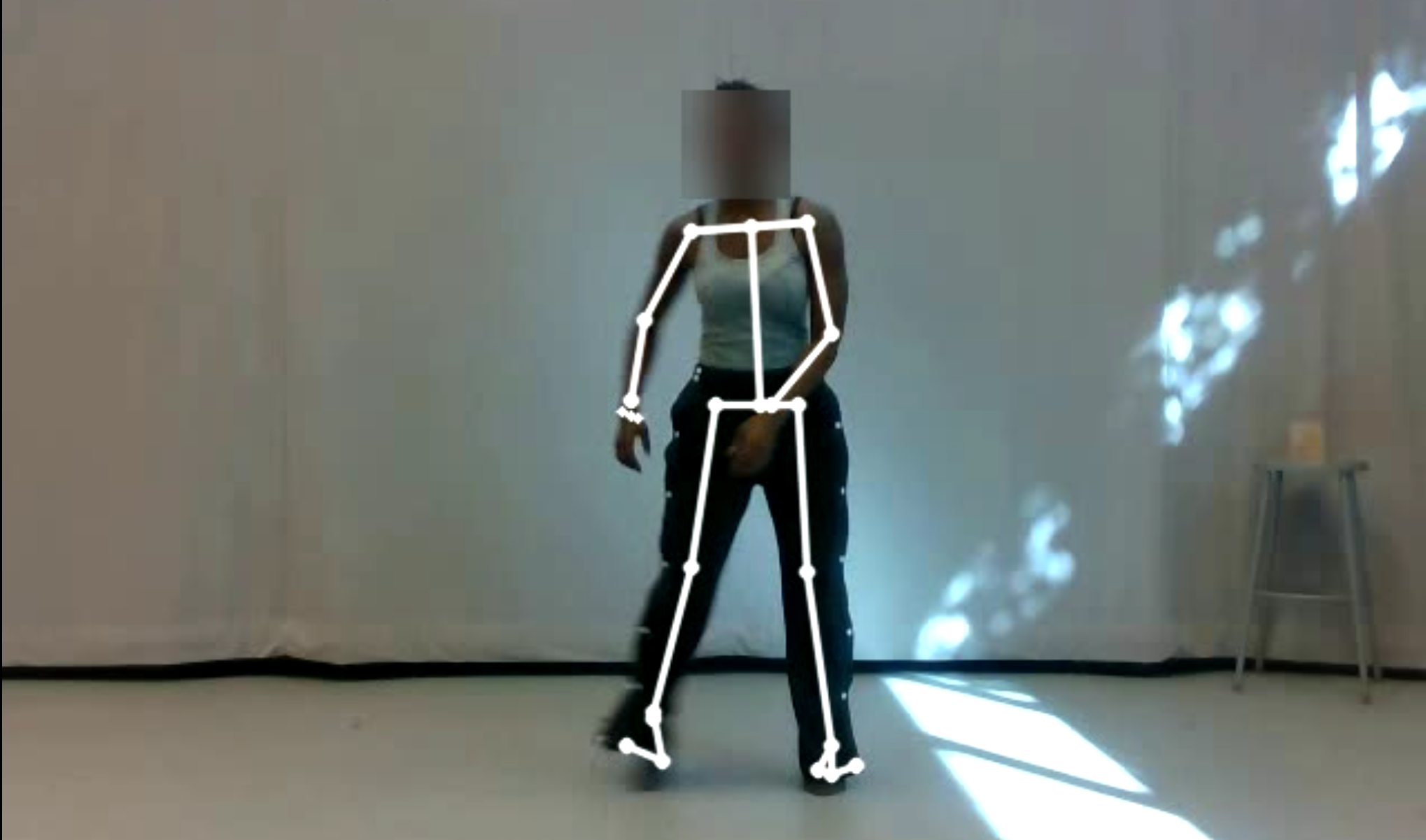} \\[2pt]

\rotatebox{90}{\scriptsize Whole-body (side)} &
\img{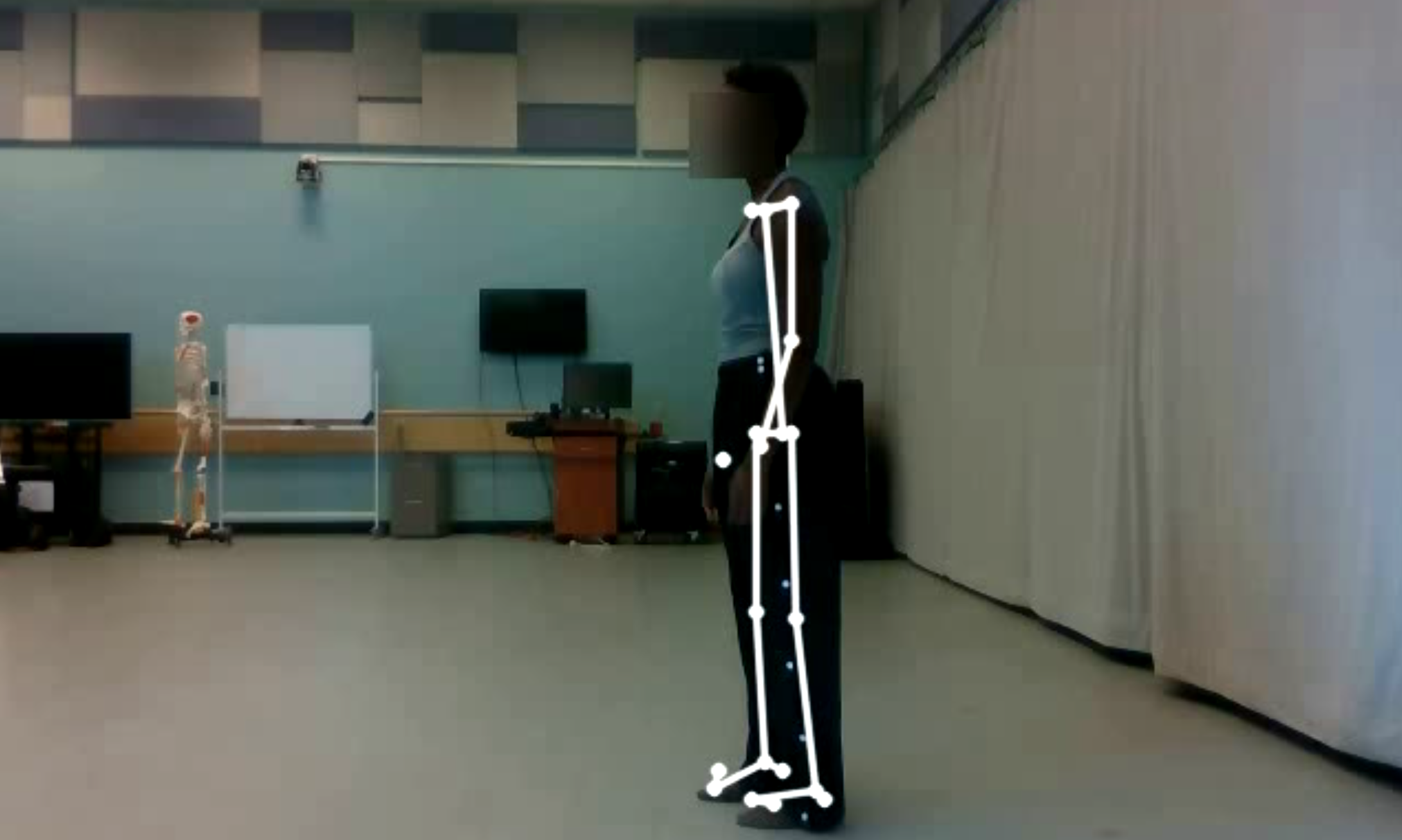} &
\img{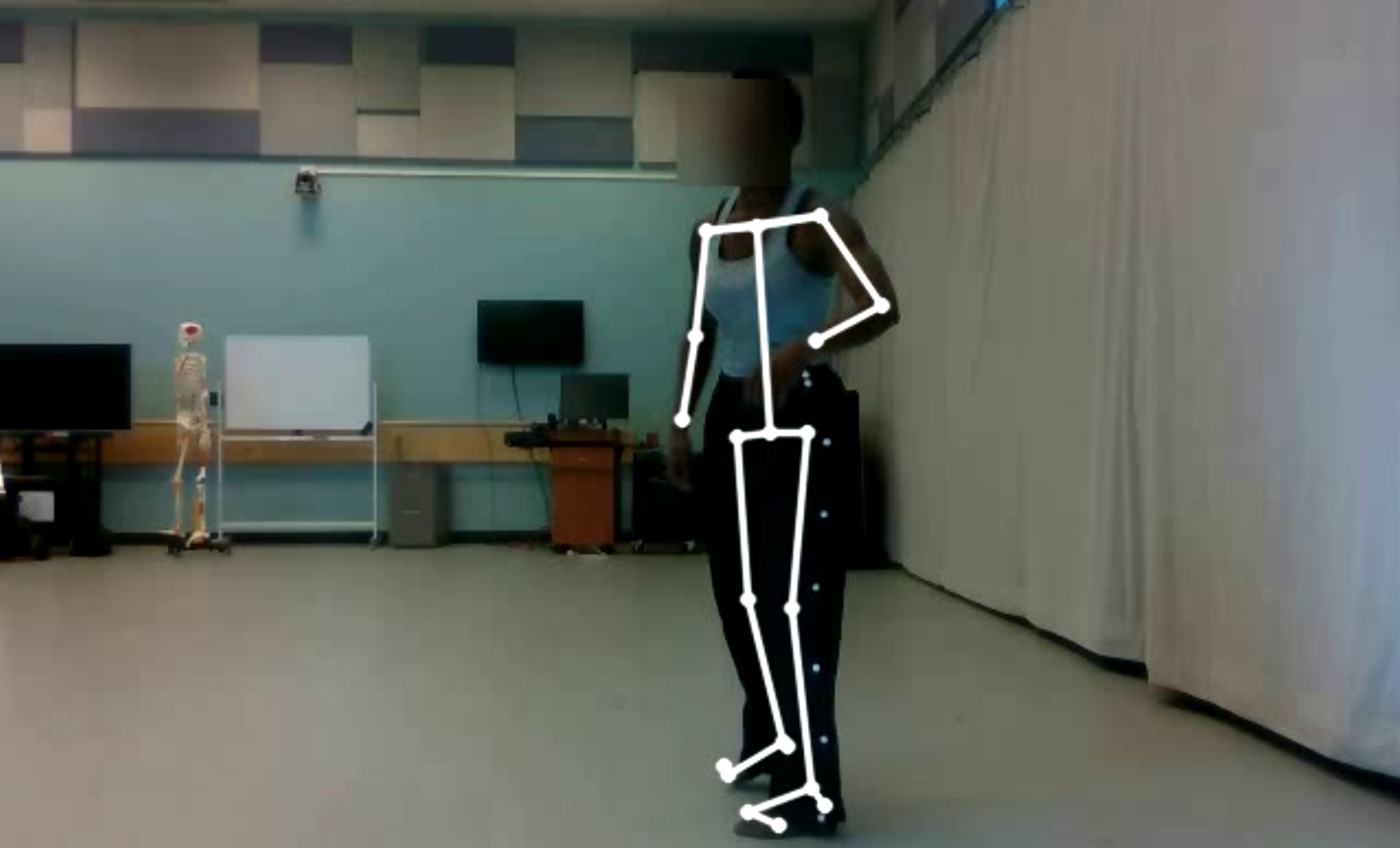} &
\img{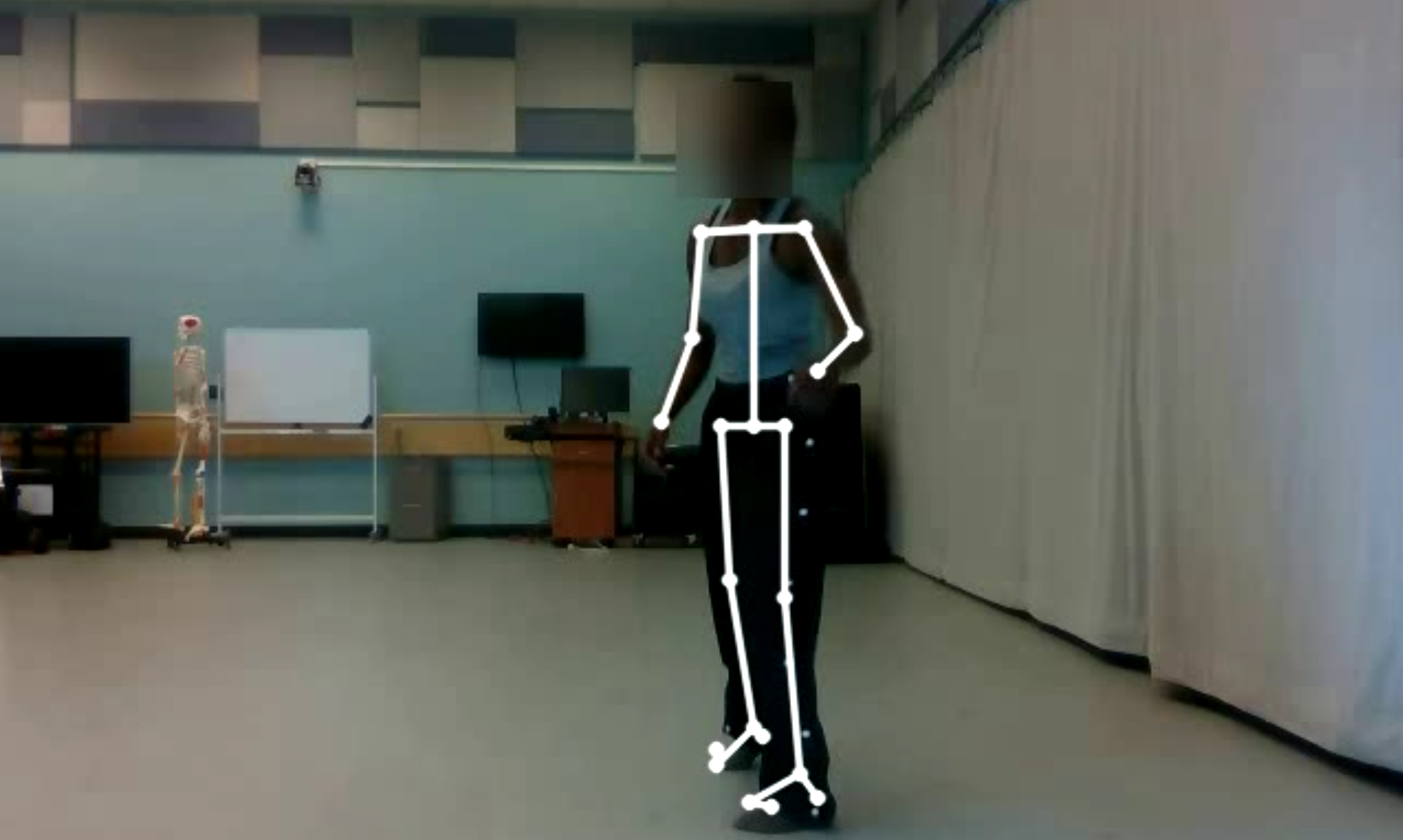} &
\img{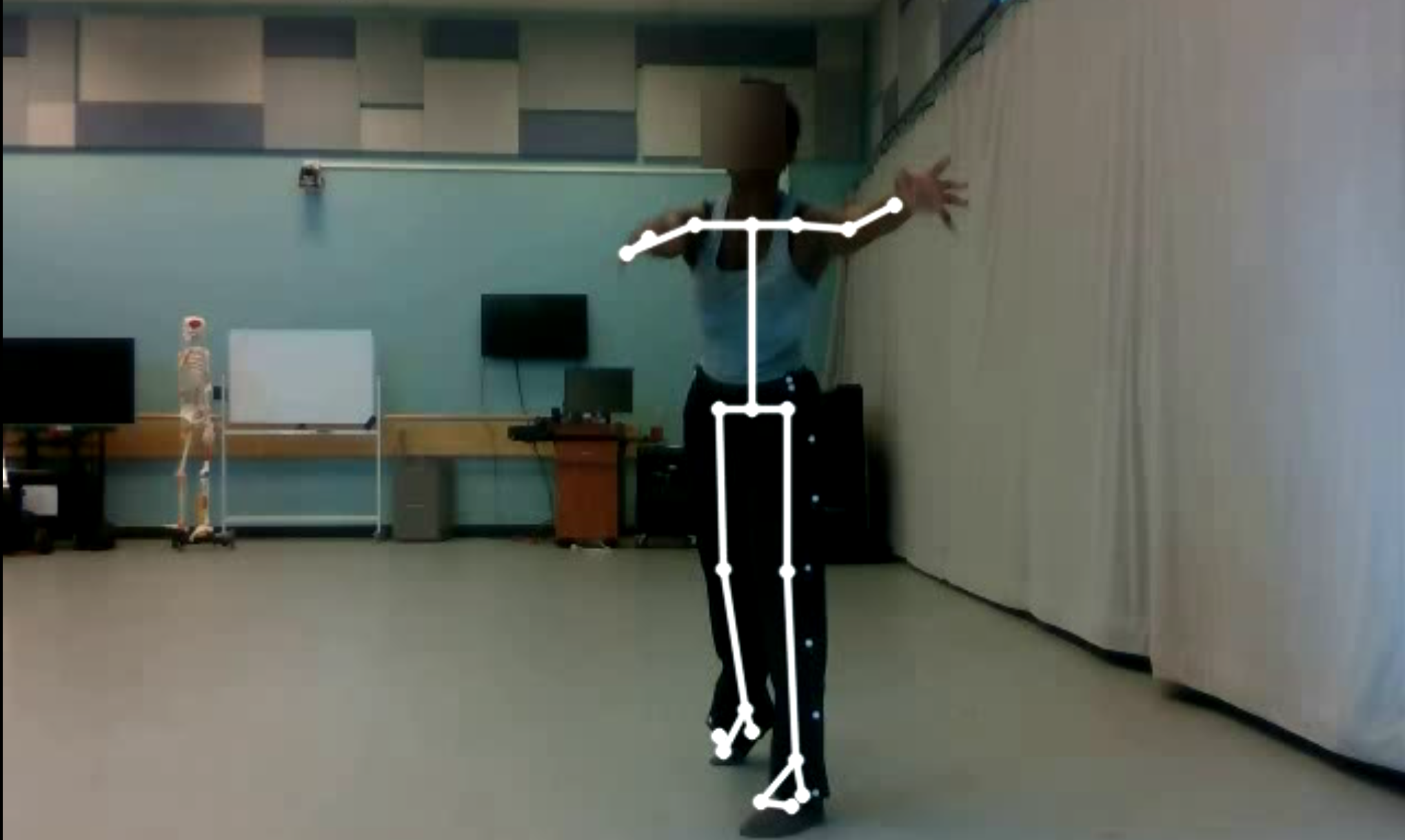} &
\img{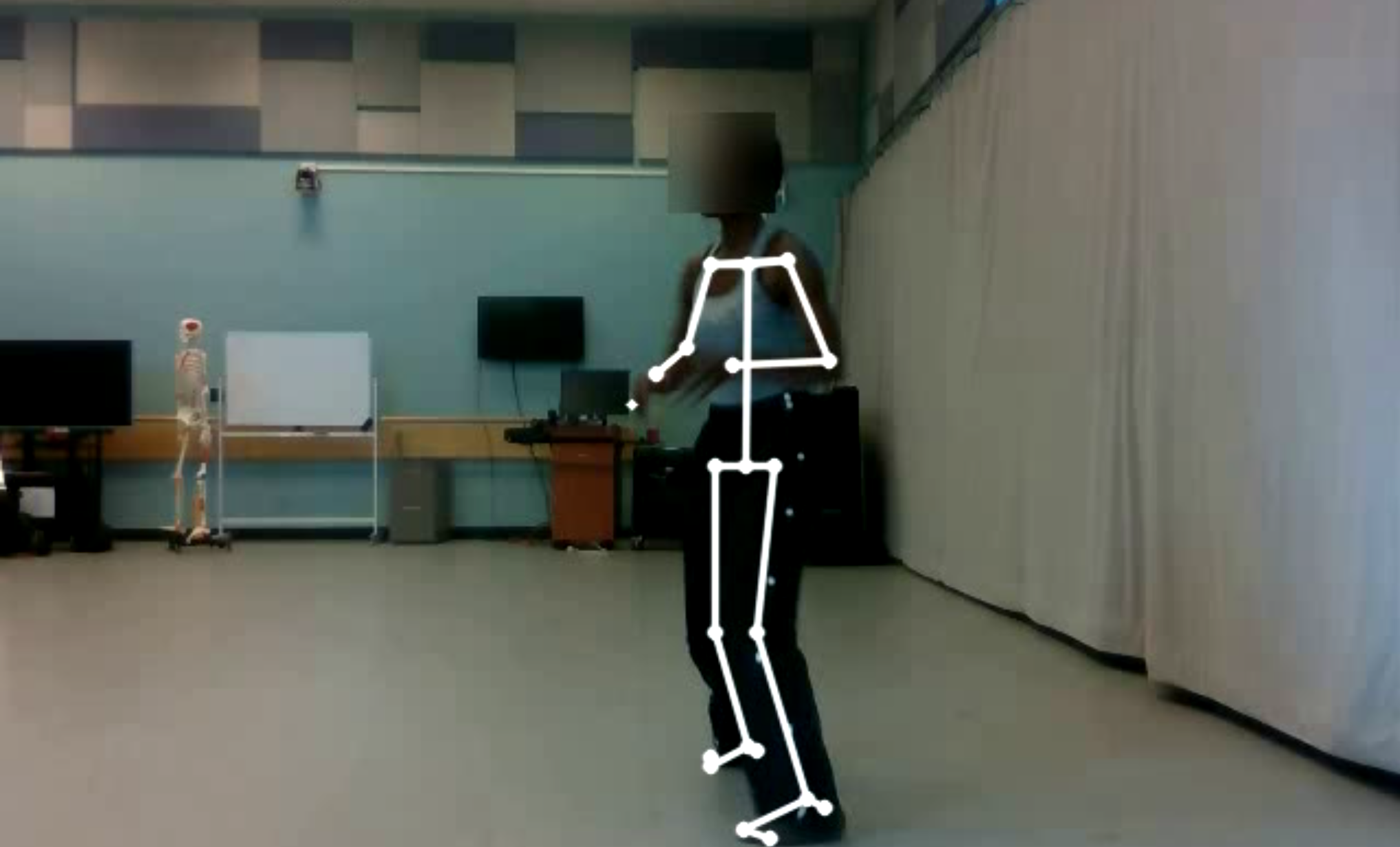} \\
\end{tabular}

\caption{Five sequential frames (1--5) from the postprocessed overlays for each view: upper-limb and whole-body motions captured from front and side cameras.}
\label{fig:sequence_grid}
\end{figure*}
\subsection{Robots expressing affect via motion}
A long-standing theme in Human-Robot interaction (HRI) is that \emph{how} a robot moves can be as communicative as \emph{what} it does \cite{dragan2013generating,fong2003socialrobots,dragan2013legibility}. Movement timing, acceleration, spatial path shape, etc., contribute to shaping human interpretations of a robot’s attitude, competence, and social stance \cite{zhou2017expressive,kulic2007physiological}. Empirical studies have shown that robot motion can measurably impact humans’ affective and subjective responses. For example, Kuli\'c and Croft study how different properties of articulated robot motion relate to physiological and subjective responses, reinforcing that motion parameters can modulate affective reactions  \cite{kulic2007physiological}. This motivated computational representations of “movement quality,” often borrowing from human movement science and performing arts \cite{knight2016laban}. A prominent example is the work proposed by Knight and Simmons, where they leverage Laban-Effort inspired motion features for mobile robots and show that such features can support classifying distinct “manners” of moving from trajectories collected with expert input, pointing toward principled feature spaces for expressivity rather than ad-hoc parameter tweaking \cite{knight2014laban}. More recent work increasingly aims to \textsl{learn} controllable affective style rather than handcraft it. One line of work treats affect as a trajectory optimization objective by learning style- or emotion-specific cost functions from user feedback and then optimizing these costs alongside task constraints \cite{zhou2018style}. A key limitation of learning separate costs for each affect label is inefficiency and poor generalization across emotions. Addressing this, Sripathy et al.\ leverage the Valence--Arousal--Dominance (VAD) affect space to learn a single mapping from trajectories to VAD, enabling a unified cost to generate motions throughout a continuous affect space \cite{sripathy2022vad}. In parallel, representation-learning approaches propose editing affective motion in a latent space. Suguitan and Hoffman introduce a classifying variational autoencoder that can modify a base affective movement toward a desired affect class, enabling post-hoc affect modulation without re-planning from scratch \cite{suguitan2020moveae}. Taken together, these works indicate that affective expression via motion is moving from handcrafted heuristics to data-driven, continuous, and controllable style representations that can be composed with task objectives.

\subsection{Robots expressing hesitancy}
Hesitancy is a particularly important subcase of expressive robot behavior because it often functions as an interactional signal. People usually express hesitation to negotiate right-of-way, communicate uncertainty, soften commitments, or coordinate in shared spaces \cite{hart2014gesture}. Moon et al. \cite{moon2013ahp}\ characterize “hesitation gestures” that occur during human-human reaching conflicts and translate them into robot motion responses for human-robot resource conflicts. They derive a characteristic Acceleration-based Hesitation Profile (AHP) from recorded human hesitation motions and evaluate whether such hesitation-like responses are recognizable and distinguishable from abrupt stopping. This formalization frames \textsl{hesitancy} as a structured, legible social cue grounded in human motion patterns.\\
Subsequent work extends hesitancy beyond reaching conflicts into a mechanism for communicating a robot’s internal uncertainty or confidence via the manner of action execution. Hough and Schlangen \cite{hough2017uncertainty} explicitly investigate how a simple robot can ground and convey different degrees of internal uncertainty through nonverbal task actions (i.e., how it acts, not just the action outcome), and show that observers can infer the robot’s internal uncertainty more reliably when the system is designed to express it in action execution. More broadly, recent work on uncertainty communication in collaborative decision-making explores embodied behaviors as one modality among others (e.g., GUIs vs.\ embodied cues) and reports that such uncertainty visualizations can meaningfully affect human decision-making and perceived transparency in high-stakes tasks \cite{schombs2024hesitancy}.\\
Taken together, prior work suggests that hesitancy is both a readable social signal and a useful mechanism for communicating uncertainty through motion. However, it also highlights a practical challenge, that expressive cues are tightly coupled to embodiment and context. This motivates a dataset that holds the functional goal constant while allowing expressive intent to vary, enabling systematic analysis of spatiotemporal signatures of hesitancy and providing training data for robot-side models. 


\section{Methods}
\label{sec:methods}
We collected a multi-modal dataset of \emph{hesitant} motion demonstrations from recruited dancers, intending to capture how hesitancy is expressed both on a robotic manipulator and through human motion. Concretely, the dataset spans:
\begin{enumerate}
    \item \textbf{Kinesthetic teaching} of a Franka Emika Panda manipulator reaching from a fixed start configuration to a fixed target (a Jenga tower), and
    \item \textbf{RGB-D motion capture} of dancer reaching motions using either the upper limb or the full body.
\end{enumerate}
Across modalities, hesitancy was recorded at three levels: \textit{slight}, \textit{significant}, and \textit{extreme}. For the full-body condition, dancers performed a single trajectory expressing \textit{extreme} hesitancy to emphasize whole-body hesitation cues.

\subsection{Robot Kinesthetic Teaching (Franka Emika Panda)}
\label{subsec:robot_kt}

\begin{figure}[h!]
    \centering
    \begin{subfigure}{\columnwidth}
  \centering
  \includegraphics[width=.8\linewidth]{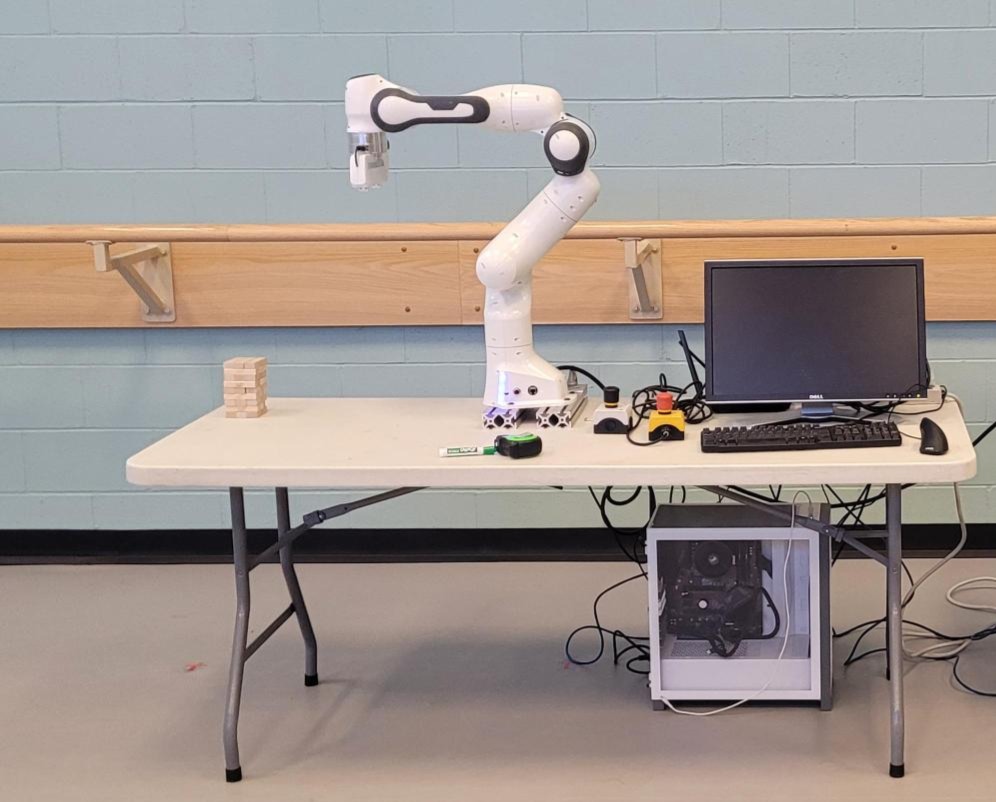}
  \caption{The kinesthetic teaching setup.}
  \label{fig:kinesthetic1}
\end{subfigure}
\begin{subfigure}{\columnwidth}
  \centering
  \includegraphics[width=.9\linewidth]{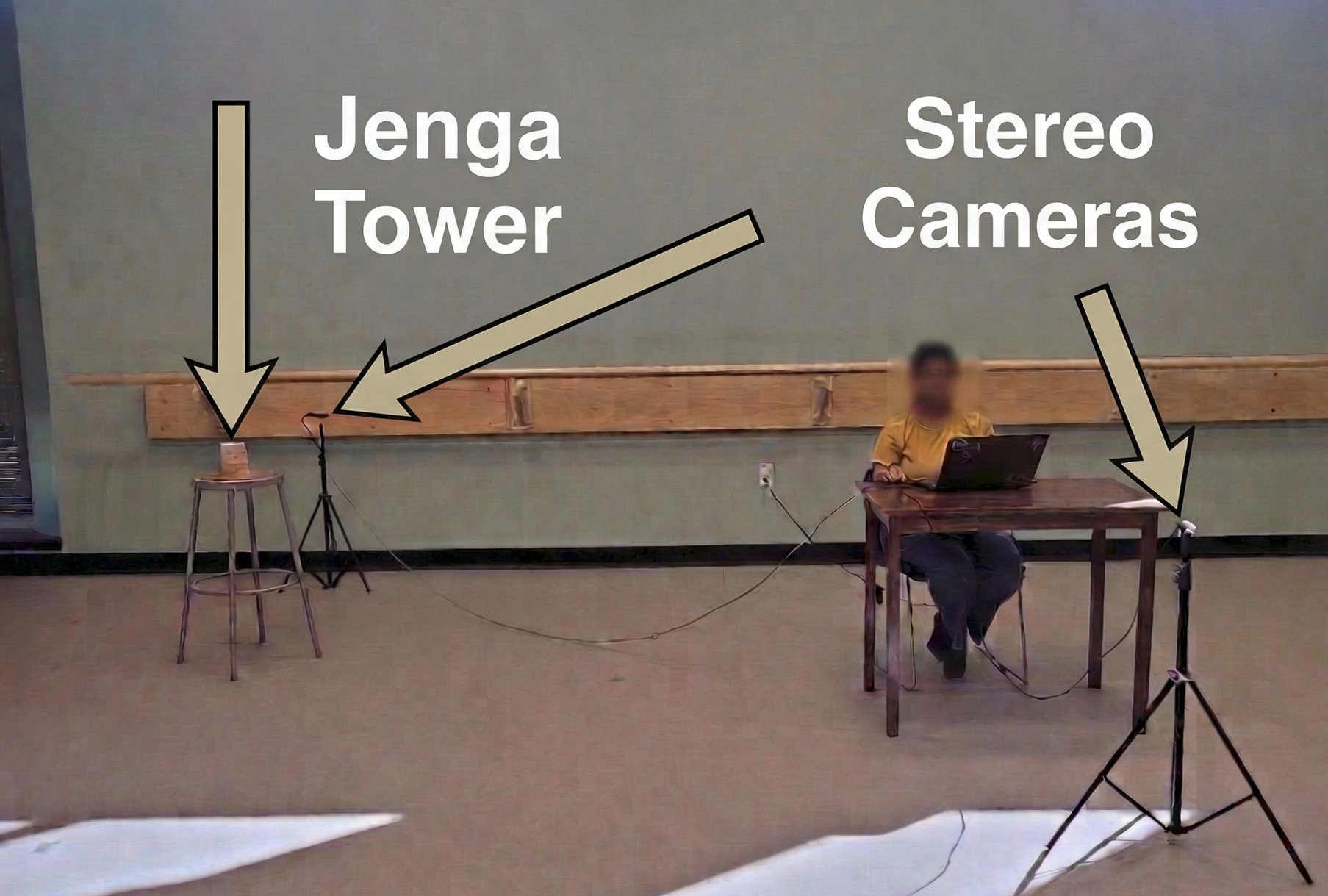}
  \caption{The motion capture setup.}
  \label{fig:mocap}
\end{subfigure}
\caption{Photos of the data collection setup featuring the kinesthetic teaching station (a) and the motion capture station (b).}
\end{figure}

We used the Franka Emika Panda robotic manipulator in a kinesthetic teaching setting (gravity-compensation / compliant guidance), where dancers physically guided the robot from its default configuration towards the Jenga tower, as seen in Fig.\ref{fig:kinesthetic} and Fig.\ref {fig:kinesthetic1}. To isolate expressive differences while holding the task goal constant, the robot always began from the same start configuration and the Jenga tower remained fixed in the workspace. Demonstrations were labeled with three hesitancy levels, and the resulting kinesthetic teaching dataset is balanced, with approximately one third of trajectories each corresponding to \textit{slight}, \textit{significant}, and \textit{extreme} hesitancy.  The dancers were allowed to watch the other demonstrations for approximately 10 minutes to familiarize themselves with the robot's motion.
During each demonstration, we logged robot measurements from the Franka ROS topic providing full robot state (\texttt{franka\_states}). For ease of access and downstream learning, we additionally export the following time-series variables to CSV:
\begin{itemize}
    \item Joint angles: $\mathbf{q}(t)$
    \item Joint velocities: $\dot{\mathbf{q}}(t)$
    \item Joint efforts/torques: $\boldsymbol{\tau}_j(t)$
    \item End-effector position: $\mathbf{p}_{ee}(t)$
    \item End-effector rotation/orientation: $\mathbf{R}_{ee}(t)$
\end{itemize}

\begin{table*}[t]
\centering
\caption{Summary of recorded modalities and file formats.}
\label{tab:data_products}
\setlength{\tabcolsep}{6pt} 
\renewcommand{\arraystretch}{0.95} 
\begin{tabularx}{\textwidth}{|p{3.3cm}|p{3.4cm}|X|p{1.5cm}|X|}
\hline
\textbf{Modality} & \textbf{Source} & \textbf{Content} & \textbf{Units} & \textbf{Stored artifacts} \\
\hline
Robot kinesthetic teaching & Franka Panda (\texttt{franka\_states}) & $\mathbf{q}, \dot{\mathbf{q}}, \boldsymbol{\tau}_j, \mathbf{p}_{ee}, \mathbf{R}_{ee}$ & SI & ROS bags; 5 CSVs (one per variable group); NPZ ($\mathbf{q}$ only) \\
\hline
Upper-limb reaching (human) & RealSense + OpenPose & 2D keypoints + confidence; raw color video & px & Videos; \texttt{keypoints\_2d.csv}; ROS bags \\
\hline
Upper-limb reaching (human) & RealSense depth & 3D keypoints + confidence; depth frames & mm & \texttt{keypoints\_3d.csv}; timestamped 16-bit depth PNG; ROS/JSON \\
\hline
Full-body extreme hesitancy & RealSense + OpenPose & BODY\_25 2D/3D keypoints + confidence; raw video & px/mm & Videos; \texttt{keypoints\_2d.csv}, \texttt{keypoints\_3d.csv}; depth PNG \\
\hline
\end{tabularx}
\end{table*}

Kinesthetic teaching data are released in three complementary formats: (i) ROS bags, (ii) five CSV files (one per variable group above), and (iii) NPZ files containing joint-angle-only trajectories $\mathbf{q}(t)$. In each CSV, \emph{columns correspond to participants} and \emph{rows correspond to timesteps}, enabling straightforward batching across demonstrations.

\subsection{Human Motion Capture with RGB-D Cameras}
\label{subsec:mocap}

To capture how dancers express hesitancy through their own movement, we recorded RGB-D motion data using two RealSense cameras (Fig.\ref{fig:mocap}), collecting synchronized color and depth streams and extracting 2D and 3D keypoints. We consider two human conditions: upper-limb reaching toward the Jenga tower (recorded at all three hesitancy levels) and a full-body sequence emphasizing \textit{extreme} hesitancy.

\subsubsection{2D keypoints (OpenPose)}
\label{subsubsec:2d}
We extract 2D keypoints using the OpenPose BODY\_25 model. Each keypoint provides pixel coordinates $(x,y)$ and a confidence score $c \in [0,1]$ per frame. To ensure consistent downstream processing, we apply a minimum confidence threshold $\texttt{MIN\_CONF}=0.30$ and treat lower-confidence detections as missing for that frame. For visualization, we provide two post-processed overlay modes:
\begin{enumerate}
    \item \textbf{Body overlay (no face):} arms, torso, legs, and feet are rendered, while face-related keypoints are not drawn.
    \item \textbf{Arm-only overlay:} left/right shoulder--elbow--wrist chains are rendered; when available, we also record hand keypoints (21 per hand) for fine-grained manipulation cues.
\end{enumerate}
In addition to MP4 videos (Fig.\ref{fig:sequence_grid}), we store raw 2D keypoints frame-by-frame as CSV files with schema \texttt{(frame, joint\_index, x, y, confidence)}, and we additionally provide equivalent streams in ROS bags for robotics toolchain compatibility.

\subsubsection{3D keypoints (Depth)}
\label{subsubsec:3d}
To obtain 3D keypoints, we align each depth frame to its corresponding color image so that the depth value at pixel $(x,y)$ matches the 2D keypoint location. Depth is recorded in millimeters. Using the camera intrinsics (focal lengths and principal point), we back-project 2D keypoints into 3D camera coordinates:
\begin{equation}
\mathbf{X} = (X,Y,Z) = \pi^{-1}(x,y,Z; \mathbf{K}),
\end{equation}
where $\mathbf{K}$ denotes intrinsics and $Z$ is the aligned depth at $(x,y)$. By default, 3D points are expressed in the camera coordinate frame (origin at the camera). When a calibrated two-camera setup is available, 3D positions may alternatively be triangulated from both views. We store the RGB-D outputs in three forms: (i) timestamped 16-bit PNG depth images (lossless), (ii) 3D keypoints as CSV with schema \texttt{(frame, joint\_index, X, Y, Z, confidence)} in millimeters, and (iii) optional JSON and ROS bag topics for 3D keypoints. Table~\ref{tab:data_products} summarizes the modalities and their stored artifacts.

For RGB-D motion capture, we target a sampling rate of 30~fps with time-aligned color, depth, and keypoints. Each depth frame and keypoint record includes timestamps to facilitate synchronization during downstream analysis. All downstream measures are computed in a confidence-aware manner: for any per-frame quantity requiring a set of keypoints (e.g., segment lengths or reach distances), we only compute values when all required joints have confidence $\ge \texttt{MIN\_CONF}$. Short confidence drops are left blank (or lightly smoothed for visualization), and we do not hallucinate or infill missing keypoints in the released raw files.

\subsection{Participants and Protocol}
We recruited $14$ dancers (2 M and 12 F, 10 dancers with 10+ years of dance training, and 4 dancers with 1-10 years of dance training). The specific genre of the dancers' expertise spanned Hip-Hop, Ballet, Contemporary, and Jazz. All 14 dancers participated in generating the motion-capture video dataset, and 11 out of 14 dancers participated in creating the kinesthetic teaching dataset. Across all dancers, we obtained 70 unique whole-body trajectories, 84 upper limb trajectories spanning over the three hesitancy levels, and 66 kinesthetic teaching trajectories spanning over the three hesitancy levels. For motion-capture visualization and analysis, we focus on body/arm keypoints and do not render facial keypoints in overlays.
\section{Conclusion and Future Work}
\label{sec:conclusion}
We presented an open-source, multi-modal dataset of \emph{hesitant} motions collected by recruiting dancers that span kinesthetic teaching demonstrations on a Franka Emika Panda and RGB-D motion capture of human reaching behaviors (upper-limb across three hesitancy levels and full-body for extreme hesitancy). By fixing the functional objective (reaching from a canonical start toward a fixed Jenga tower) and varying only the intended hesitancy, the robot and upper limb dataset isolates hesitancy as a form of \emph{functional expressivity} expressed through spatiotemporal motion cues.
Looking forward, this dataset enables several research directions.\\
On the robotics side, kinesthetic teaching trajectories can support learning hesitancy-aware representations and controllers, including supervised recognition of hesitancy from robot state, as well as conditional generative modeling (e.g., diffusion-based synthesis) to produce a continuous ``hesitancy slider'' that interpolates between hesitancy levels while satisfying task constraints. More broadly, hesitant motion planning can be framed as optimizing feasibility and safety while regularizing trajectories toward a learned hesitancy prior. On the human side, the RGB-D recordings enable benchmark tasks for hesitancy recognition from 2D keypoints, 3D keypoints, and RGB, as well as analyses of which kinematic features most strongly correlate with perceived hesitancy. Finally, the dataset provides a foundation for cross-embodiment modeling.
Future work can investigate shared latent embeddings of hesitancy that align human and robot trajectories without explicit retargeting, and can extend whole-body hesitant behaviors to humanoids. As an additional direction for future work, it would be valuable to study how experts from different communities express hesitancy under the same task constraints. In particular, comparing demonstrations from dancers with those from puppeteers, actors, and other performance practitioners could reveal which kinematic cues are consistent across training backgrounds. Such comparisons may help disentangle ``universal'' hesitancy signatures (e.g., pauses, delayed commitment) from domain-specific embellishments, and inform data collection strategies that better balance expressivity, realism, and generalization to everyday users. We hope this release accelerates research on motion as a mode of communication by providing reproducible resources for studying, modeling, and generating hesitancy as a task-relevant nonverbal signal.

\section{Acknowledgments}

We would like to thank Prof. Madeline Harvey at Colorado State University for her support and time. We thank the dancers at Colorado State University for their time and creativity, leading to the creation of this dataset. 

\bibliographystyle{ACM-Reference-Format}
\bibliography{software}
\end{document}